\documentclass[10pt,twocolumn,letterpaper]{article}

\usepackage{multirow}
\usepackage{multicol}
\usepackage{arydshln}
\usepackage{colortbl}
\usepackage{booktabs}
\usepackage{float}

\newcommand{\blue}[1]{\textcolor{blue}{#1}}
\usepackage{cvpr}              

\newcommand{\red}[1]{{\color{red}#1}}

\definecolor{cvprblue}{rgb}{0.21,0.49,0.74}
\usepackage[pagebackref,breaklinks,colorlinks,allcolors=cvprblue]{hyperref}

\def\paperID{2} 
\def\confName{CVPR}
\def\confYear{2025}

\title{\includegraphics[width=7mm]{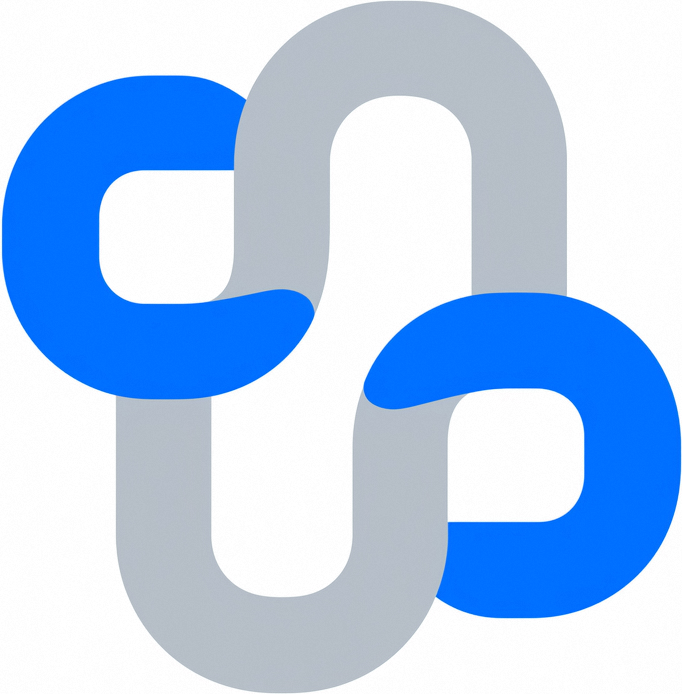} UniToken: Harmonizing Multimodal Understanding and Generation through Unified Visual Encoding}

\author{
Yang Jiao\textsuperscript{1,2*},
Haibo Qiu\textsuperscript{3}\thanks{Equal contribution. $^{\dagger}$ Corresponding authors.},
Zequn Jie$^{\dagger}$\textsuperscript{3},
Shaoxiang Chen\textsuperscript{3},
Jingjing Chen$^{\dagger}$\textsuperscript{1,2},\\
Lin Ma\textsuperscript{3}, and
Yu-Gang Jiang\textsuperscript{1,2}\\
\textsuperscript{1} Shanghai Key Lab of Intell. Info. Processing, School of CS, Fudan University \\
\textsuperscript{2} Shanghai Collaborative Innovation Center on Intelligent Visual Computing \\
\textsuperscript{3} Meituan \\
{\tt \small yjiao23@m.fudan.edu.cn, }
{\tt \small haibo-qiu@outlook.com} \\
{\tt \small \{zequn.nus, forest.linma\}@gmail.com, }
{\tt \small \{sxchen13, chenjingjing, ygj\}@fudan.edu.cn}
}

\begin{document}
\maketitle
\begin{abstract}
   We introduce UniToken, an auto-regressive generation model that encodes visual inputs through a combination of discrete and continuous representations, enabling seamless integration of unified visual understanding and image generation tasks. Unlike previous approaches that rely on unilateral visual representations, our unified visual encoding framework captures both high-level semantics and low-level details, delivering multidimensional information that empowers heterogeneous tasks to selectively assimilate domain-specific knowledge based on their inherent characteristics. Through in-depth experiments, we uncover key principles for developing a unified model capable of both visual understanding and image generation. Extensive evaluations across a diverse range of prominent benchmarks demonstrate that UniToken achieves state-of-the-art performance, surpassing existing approaches. These results establish UniToken as a robust foundation for future research in this domain. The code and models are available at \hyperlink{https://github.com/SxJyJay/UniToken}{https://github.com/SxJyJay/UniToken}.
   
\end{abstract}

\section{Introduction}
With next-token prediction demonstrating promising scaling laws in both language modeling~\cite{touvron2023llama,achiam2023gpt,bai2023qwen} and multimodal comprehension~\cite{liu2024visual,bai2023qwenvl,jiao2024lumen,zhang2024eventhallusion,li2024look,hong2025worldsense}, exploring similar scaling potential in the field of visual generation~\cite{sun2024autoregressive,wang2024omnitokenizer} has become a recent trend. LlamaGen~\cite{sun2024autoregressive} and OmniTokenizer~\cite{wang2024omnitokenizer} demonstrate the scalability of transformer architectures in the domains of image and video generation, respectively. Sharing similar model architectures and learning paradigms, the unification of multimodal understanding and generation within a shared model offers promising prospects and has recently garnered significant research interest~\cite{lu2024chameleon,wang2024emu3,wu2024vila,zhou2024transfusion,wu2024janus}. Existing works can be categorized into two groups based on their visual encoding paradigms. The first group, exemplified by Chameleon~\cite{lu2024chameleon} and Emu3~\cite{wang2024emu3}, encodes visual inputs into discrete tokens predefined within a fixed vocabulary. The second group, represented by Unified-IO2~\cite{lu2024unified} and Janus~\cite{wu2024janus}, adopts a decoupled visual encoding approach, utilizing continuous visual tokens for image comprehension and discrete tokens for image generation.

However, these two paradigms face challenges stemming from their respective visual encoding techniques. For the discrete-only encoding paradigm, the efficiency of absorbing knowledge from multimodal data lags significantly behind that of its continuous-only counterpart, as proved in Tab.\ref{tab:task_interference}. This phenomenon arises from the information loss incurred during the quantization process used to generate discrete tokens. Moreover, discarding the continuous visual tokens sacrifices the sweetness of advanced image encoding techniques~\cite{liu2024improved,luo2024mono}. 
On the other hand, in the decoupled encoding paradigm, employing different visual encoders for distinct tasks restricts the model's flexibility, and the burdens introduced by such mode switching increase as the model's functionality continues to scale up.

\begin{figure*}[!t]
  \centering 
  \includegraphics[width=0.8\textwidth]{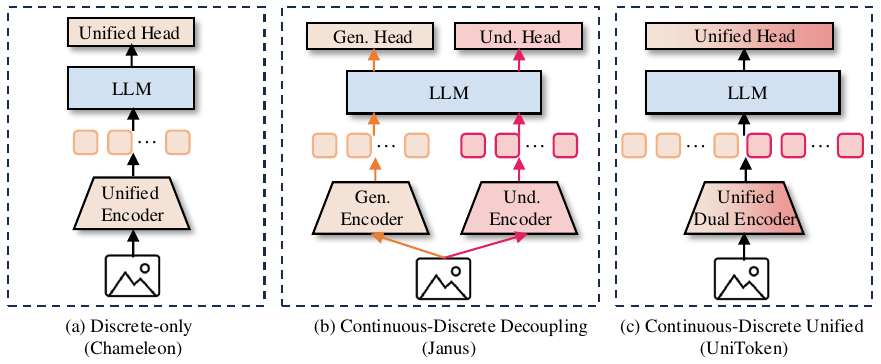} 
  \caption{\textbf{Different visual encoding paradigms for developing a unified model for visual understanding and image generation.} We use \textcolor{orange}{orange} to denote components related to discrete visual encoding and \red{red} to signify components associated with continuous visual encoding.  For the sake of brevity, we omit the text tokens in the image.} 
  \label{fig:paradigm_cmp} 
\end{figure*}

Toward this end, in this paper, we propose UniToken, which employs a unified visual encoding that is agnostic to specific tasks. As shown in Fig.\ref{fig:paradigm_cmp}, we encode an image with dual visual encoders, resulting in a combination of discrete and continuous tokens that form the unified visual representation. Carrying both low-level details and high-level semantics, this unified visual representation can seamlessly support both image understanding and generation tasks. Additionally, we employ several advanced techniques, namely scaling up resolutions and tuning the ViT, to significantly amplify the visual representation. By incorporating these techniques, multimodal comprehension capabilities, especially in text-reading-related scenarios, are significantly boosted. 

Through extensive experiments, we uncover several valuable empirical insights for developing a unified multimodal understanding and generation model: (1) \textbf{Task interference:} a combined discrete and continuous visual encoding ensures seamless compatibility between multimodal understanding and generation tasks, while relying solely on discrete visual tokens is prone to task interference.  (2) \textbf{Task-specific data distribution:} the data proportion between multimodal understanding and generation tasks should be adjusted when using training data of varying scales.  Furthermore, through systematic evaluation across a wide array of prevalent visual understanding and image generation benchmarks, our UniToken outperforms unified models for both understanding and generation and competes with state-of-the-art specialists in each respective field.

In summary, our contributions are three-fold: 
\begin{itemize}
    \item We propose UniToken, which integrates both discrete and continuous tokens as a unified visual representation to deal with diverse tasks.
    \item We further equip our UniToken with advanced visual encoding techniques, significantly boosting our model's comprehension capabilities.
    \item We uncover several empirical insights of developing a unified model for multimodal understanding and generation. And our UniToken can compete with state-of-the-art approaches in both multimodal understanding and generation fields.
\end{itemize}

\section{Related Works}
\subsection{Multimodal Understanding}
Building upon powerful Large Language Models (LLMs), MLLMs demonstrate impressive multimodal content comprehension and reasoning capabilities. To efficiently adapt visual concepts to the textual world of LLMs, mainstream MLLMs~\cite{liu2024visual,bai2023qwenvl,chen2025sharegpt4v,jiao2024lumen} leverage vision encoders such as CLIP~\cite{radford2021learning} or SigLIP~\cite{zhai2023sigmoid}, which are enriched with semantics derived from vision-text contrastive training. By further tuning an adapter and the entire model sequentially with multimodal instructional data, knowledge encapsulated in pretrained LLMs can effectively fuel the comprehension of the visual world. However, the visual features from these vision encoders inherently lack the image generation potential. To mitigate this gap, recent advancements~\cite{ge2023planting,ge2024seed,jin2023unified,dong2023dreamllm} integrate an external diffusion model with the MLLM, leveraging MLLM outputs to generate the conditions for the diffusion process. The generative capability of this paradigm may be limited by the external diffusion model and the effectiveness of adapting the MLLM's outputs to the diffusion conditional space.

\subsection{Visual Generation}
To explore the scaling property in the visual generation field, attention has been switched to the autoregressive-based paradigm. VQ-VAE~\cite{van2017neural} and VQ-GAN~\cite{esser2021taming} are pioneers of this paradigm, where the image is first quantized into a sequence of discrete codebook IDs, and then their underlying distribution is modeled with transformer architectures. To reduce the quantization error, recent works~\cite{lee2022autoregressive,yu2023language,luo2024open} develop advanced tokenization techniques and explore the trade-off between codebook size and codebook embedding dimension. Building on these powerful image tokenizers, recent advancements~\cite{sun2024autoregressive,wang2024omnitokenizer} have scaled up training data and transformer parameters, revealing scaling properties similar to those observed in the LLM field. However, as the tokens within the codebook are designed primarily for image reconstruction and generation tasks, they predominantly capture low-level visual details while overlooking high-level semantics, posing significant challenges for extending their application to the multimodal comprehension domain.

\begin{figure}[!t]
  \centering 
  \includegraphics[width=0.5\textwidth]{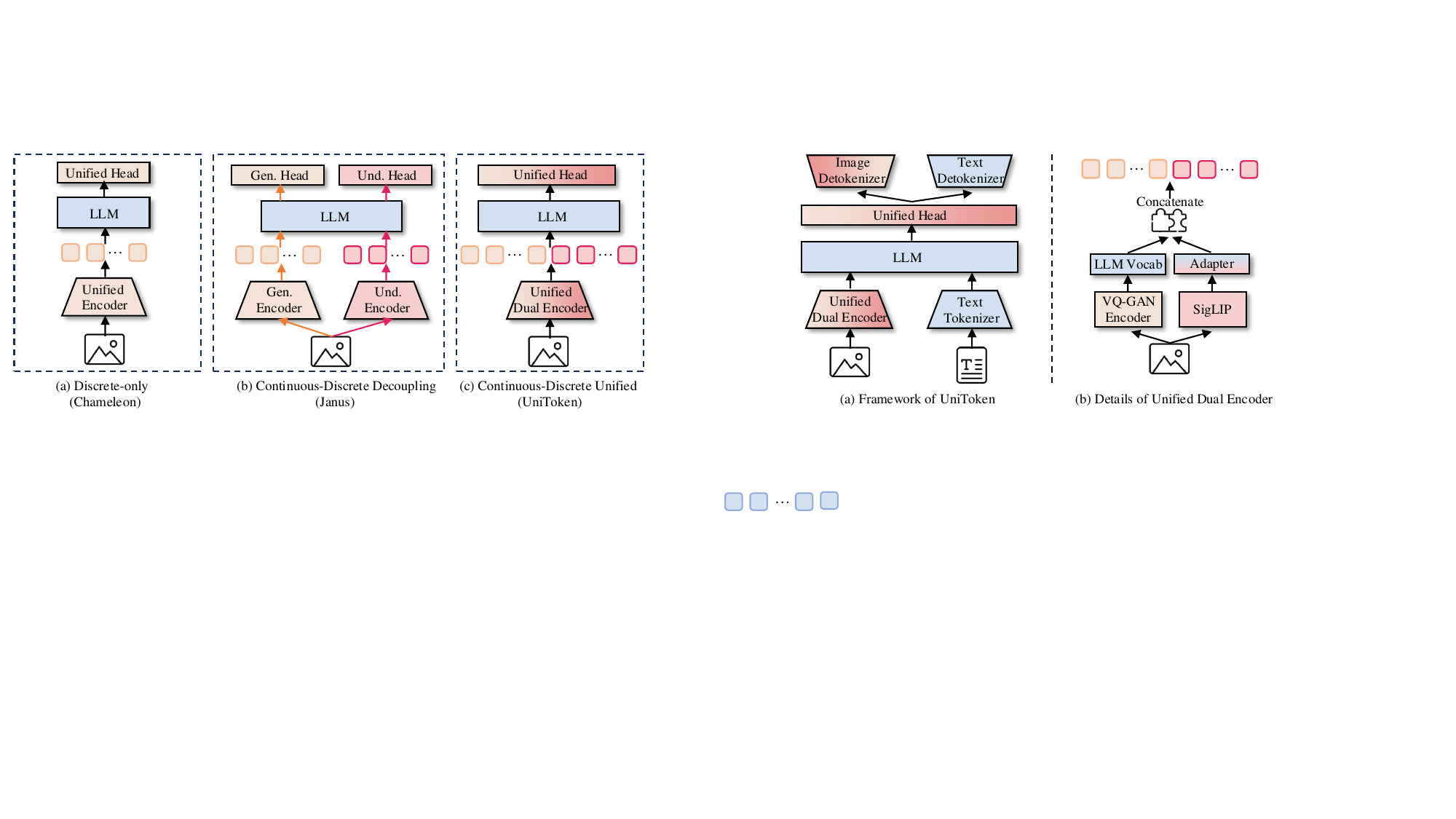} 
  \caption{Illustration of \textbf{(a) the overall framework of UniToken} and \textbf{(b) detailed designs of the unified dual encoder presented in (a)}. In (a), the ``image detokenizer" and ``text detokenizer" are responsible for converting predicted token IDs back into images and words, respectively. In (b), the VQ-GAN encoder processes an image and outputs discretized token IDs, which are then transformed into high-dimensional embeddings by indexing the LLM's vocabulary.} 
  \label{fig:unitoken_framewark} 
  \vspace{-8pt}
\end{figure}

\subsection{Unified Understanding and Generation}
Unifying multimodal understanding and generation within a shared model facilitates the seamless association between high-level content reasoning with low-level image generation capabilities, allowing them to complement and enhance each other. Chameleon~\cite{lu2024chameleon} and Lumina-mGPT~\cite{liu2024lumina} employ the VQ-GAN's image tokenizer as the visual encoder, and jointly train the LLM with VQA and image generation data. However, their multimodal understanding performances lag significantly behind conventional MLLMs like LLaVA-v1.5~\cite{liu2024improved} and Qwen-VL~\cite{bai2023qwenvl}. To address this, VILA-U~\cite{wu2024vila} distills high-level semantics from CLIP when training its own image tokenizer. Emu3~\cite{wang2024emu3} also trains a private image tokenizer based on MoVQGAN~\cite{zheng2022movq}, which has a lower compression rate (8$\times$8) compared with previous tokenizers. Recently, Janus~\cite{wu2024janus} tackles this problem with a more straightforward design, which decouples the visual encoding of understanding and generation with the image tokenizer from SigLIP~\cite{zhai2023sigmoid} and LlamaGen~\cite{sun2024autoregressive}, respectively. Meanwhile, it employs two decoupled prediction heads for the two tasks. While effective, this decoupled design imposes a significant mode-switching burden when scaling to more complex tasks. In contrast, our UniToken adopts a unified visual encoding and prediction head, irrespective of task types.

\section{UniToken}
\subsection{Architectural Designs}
The overall framework of UniToken is illustrated in Fig.~\ref{fig:unitoken_framewark}(a). Given a multimodal input sequence, textual inputs are tokenized using the LLM's tokenizer, while visual inputs are encoded through a dual visual encoder to produce a unified representation of continuous and discrete visual tokens. These tokens are subsequently fed into the LLM, which generates output tokens corresponding to either images or text. Finally, the output tokens are processed by their respective de-tokenizers to produce the final results. In the following section, we will elaborate on the designs of unified visual encoding and the advanced techniques, both of which are essential for developing a robust unified multimodal understanding and generation model.

\begin{table*}[!t]
\begin{center}
\scalebox{0.9}{
\begin{tabular}{c|c|c}
\toprule
Task                         & Type           & Dataset                                                        \\ \midrule
\multirow{4}{*}{Visual Und.} & Caption        & ShareGPT4V(200K)~\cite{chen2025sharegpt4v}, SA1B-DenseCap(100K)~\cite{chen2024far}                        \\
                             & GeneralQA      & LLaVA-1.6(1.5M)~\cite{liu2024improved}, M3IT(920K)~\cite{li2023m3it}, ALLaVA(1.35M)~\cite{chen2024allava}                  \\
                             & Doc/Chart/Math & ChartQA(390K)~\cite{masry2022chartqa}, UReader(1.1M)~\cite{ye2023ureader}, ScigraphQA(296K)~\cite{li2023scigraphqa}, Geo170K(118K)~\cite{gao2023g} \\
                             & OCR            & TextOCR(25K)~\cite{singh2021textocr}, PDFA(3M)~\cite{pdfa}, SynthDoG(800K)~\cite{kim2022ocr}, WebSRC(100K)~\cite{chen2021websrc}       \\ \midrule
Image Gen.                   & Text2Img Gen.  & MidJouney(10M)                                                \\ \bottomrule
\end{tabular}}
\end{center}
\caption{\textbf{Distribution of training data in Stage II.} Except ``MidJorney(10M)", all datasets are publicly available for academic purpose.}
\label{tab:data_dist.}
\end{table*}

\vspace{1.5mm}
\noindent\textbf{Unified Visual Encoding.}\quad As illustrated in the Fig.\ref{fig:unitoken_framewark}(b), the dual visual encoder comprises a SigLIP~\cite{zhai2023sigmoid} and a VQ-Tokenizer~\cite{esser2021taming} from Chameleon~\cite{lu2024chameleon}. SigLIP ViT extracts continuous image features that are semantically rich, which are then aligned with the LLM's input space via a two-layer MLP. Meanwhile, the VQ-Tokenizer discretizes an image into a sequence of codebook IDs, which are subsequently used to retrieve their corresponding high-dimensional features from the LLM's vocabulary. Combining the continuous and discrete image features, a multimodal input sequence can be formulated as follows:
\begin{equation}
    \mathtt{[BOS]}\mathtt{[BOI]}\{\mathrm{image\_d}\}\mathtt{[SEP]}\{\mathrm{image\_c}\}\mathtt{[EOI]}\{\mathrm{text}\}\mathtt{[EOS]} \notag
\end{equation}
where $\mathtt{[BOS]}$ and $\mathtt{[EOS]}$ denote special tokens that mark the beginning and end of the entire sequence, respectively, while $\mathtt{[BOI]}$ and $\mathtt{[EOI]}$ represent special tokens that indicate the start and end of the image segment. Additionally, $\mathtt{[SEP]}$ serves as a special token for separating the discrete ($\{\mathrm{image\_d}\}$) and continuous image features ($\{\mathrm{image\_c}\}$). Carrying both high-level semantics and low-level details of the input image, the unified visual encoding facilitates the LLM to selectively assimilate knowledge to deal with different tasks.

\vspace{1.5mm}
\noindent\textbf{Advanced Techniques.}\quad To further enhance visual knowledge, we integrate two state-of-the-art off-the-shelf techniques to augment the unified visual encoding features. (1) To scale the continuous representation to a higher resolution, we partition the image into multiple grids and encode each grid independently as done in~\cite{liu2024improved}. We choose the grid configuration of $\{2\times2,1\times\{2,3\},\{2,3\}\times1\} $ to support image inputs of varying shapes. (2) To dynamically adjust continuous representation end-to-end, we finetune the SigLIP ViT throughout our training process. Empirically, we observe that a large learning rate causes the ViT to collapse. To address this, we carefully regulate the learning rate magnitude for the ViT, resulting in substantial performance improvements across a wide range of benchmarks.

\subsection{Training Recipes}  
To effectively harmonize visual understanding and image generation capabilities, we train UniToken using a two-stage pipeline. In the following sections, we provide a detailed explanation of the training procedures for each stage.

\vspace{1.5mm}
\noindent\textbf{Stage I:} We utilize Chameleon~\cite{lu2024chameleon} as our foundation model, which inherently supports image discretization and the distribution modeling of these discretized image tokens. To align the continuous image features with the base model, we freeze the LLM parameters and train only the SigLIP ViT and adapter components during Stage I. During Stage I, we utilize a dataset of 2.5 million image captions, composed of data from ShareGPT4V ($49.5\%$), LLaVA ($22.2\%$), and ALLaVA ($28.2\%$).  

\vspace{1.5mm}
\noindent\textbf{Stage II:} This stage aims to develop both visual understanding and image generation capabilities for the LLM with the aid of the unified visual encoding features. Toward this end, we train all parameters, including ViT, adapter and the LLM, during Stage II. As depicted in Tab.\ref{tab:data_dist.}, in Stage II, we utilize a dataset comprising 10 million image-to-text understanding samples and 10 million text-to-image generation samples, totaling 20 million data points. The image-to-text understanding dataset encompasses image captions, general question-answer pairs, documents, charts, mathematical problems, and OCR instances, all of which are publicly available. Additionally, the text-to-image generation data are manually synthesized by prompting the Midjourney model. The ratio between multimodal understanding and generation data is approximately 1:1. We further investigate the impact of this ratio across different data scales in Sec.~\ref{sec_exp:data_distribution}.

\vspace{1.5mm}
\noindent\textbf{Stage III:} This stage aims to further enhance the instruction-following capabilities for both visual understanding and image generation. We adopt the same training strategy as Stage II while curating exceptionally high-quality multimodal conversation data (423K data points) and text-to-image generation data (100K data points). The multimodal conversation data encompasses long-context dialogues set in realistic scenarios and documents with diverse content. The text-to-image data is carefully curated to specifically enhance object-centric control. Through Stage III training, UniToken achieves substantial performance improvements in OCR-oriented benchmarks and text-to-image generation precision.

\subsection{Training Objectives}
Following the prior auto-regressive prediction-based approaches, we adopt cross-entropy as the loss function:
\begin{equation}
    \mathcal{L}=-\sum_{i}\mathrm{log}P_{\theta}(x_i|x_{<i})
\end{equation}
where $\theta$ denotes the model parameters. For the visual understanding task, we compute the loss exclusively over the text tokens corresponding to the answer. For the image generation task, we calculate the loss solely over the discrete image tokens.

\begin{table}[!t]
\begin{center}
\scalebox{0.8}{
\begin{tabular}{l|cc}
\toprule
                     & Stage I                  & Stage II,III                  \\ \midrule
Global Learning Rate & $1e^{-3}$                     & $1e^{-5}$                      \\
ViT Learning Rate    & $1e^{-5}$                     & $1e^{-5}$                      \\
LR Scheduler         & Cosine                   & Cosine                    \\
Optimizer            & \multicolumn{2}{c}{AdamW($\beta_{1}=0.9, \beta_{2}=0.95$)} \\
Batch Size           & 128                      & 128                       \\
Epoch                & 1                        & 1                         \\ \bottomrule
\end{tabular}}
\end{center}
\caption{\textbf{Details of training hyperparameters}. ``Global Learning Rate" is applied to all model parameters except for the ViT.}
\label{tab:hyperparam}
\end{table}

\begin{table*}[!t]
\begin{center}
\scalebox{0.8}{
\begin{tabular}{lcccccccccc}
\toprule
\multicolumn{1}{l|}{Method}           & \multicolumn{1}{c|}{LLM}              & MMMU          & MMB-CN        & MMB-EN        & MMStar        & SEED          & Math-Vista    & HBench        & AI2D          & OCRBench               \\ \midrule
\multicolumn{11}{c}{\textit{Understanding-Only}}                                                                                                                                                                                                      \\ \midrule
\multicolumn{1}{l|}{InstructBLIP~\cite{dai2023instructblip}}     & \multicolumn{1}{c|}{Vicuna-7B}        & 30.6          & 23.9          & 36.0          & -             & 53.4          & 25.3          & 45.3          & -             & 276                   \\
\multicolumn{1}{l|}{QwenVL-Chat~\cite{bai2023qwenvl}}      & \multicolumn{1}{c|}{Qwen-7B}          & 35.9          & 56.3          & 60.6          & \textbf{37.5} & 65.4          & -             & 39.2          & 45.9          & 488                       \\
\multicolumn{1}{l|}{mPLUG-Owl2~\cite{ye2024mplug}}       & \multicolumn{1}{c|}{LLaMA2-7B}        & 32.7          & 60.7          & 64.5          & -             & 57.8          & 22.2          & -             & -             & 255                    \\
\multicolumn{1}{l|}{LLaVA-v1.5~\cite{liu2024improved}}       & \multicolumn{1}{c|}{Vicuna-7B}        & 35.3          & 46.4          & 64.3          & 30.3          & 64.3          & -             & \textbf{46.9} & 54.8          & 318                    \\
\multicolumn{1}{l|}{ShareGPT4V~\cite{chen2025sharegpt4v}}       & \multicolumn{1}{c|}{Vicuna-7B}        & \textbf{37.2} & 60.7          & \textbf{68.8} & 33.0          & -             & -             & -             & \textbf{58.0} & 371                    \\
\multicolumn{1}{l|}{DeepSeek-VL~\cite{lu2024deepseek}} & \multicolumn{1}{c|}{DeepSeek-1B}  & 32.2          & \textbf{62.9} & 64.6          & -             & \textbf{66.7} & -             & 27.6          & -             & 409                    \\
\multicolumn{1}{l|}{LLaVA-v1.6(HD)~\cite{liu2024llava}}   & \multicolumn{1}{c|}{Vicuna-7B}        & 35.1          & 62.3          & 67.4          & -             & 64.7          & \textbf{34.6} & -             & 66.6*         & \textbf{532}  \\ \midrule
\multicolumn{11}{c}{\textit{Understanding \& Generation}}                                                                                                                                                                                                   \\ \midrule
\multicolumn{1}{l|}{Chameleon~\cite{lu2024chameleon}}        & \multicolumn{1}{c|}{-}                & \textbf{34.4} & 26.0          & 31.6          & 30.3          & 47.4          & 13.6          & 17.8          & 45.6          & 18                     \\
\multicolumn{1}{l|}{Lumina-mGPT~\cite{liu2024lumina}}      & \multicolumn{1}{c|}{Chameleon7B}      & 25.1          & 16.1          & 33.0          & 28.2          & 51.3          & 21.2          & 35.8          & 44.1          & 37                  \\
\multicolumn{1}{l|}{Show-o~\cite{xie2024show}}           & \multicolumn{1}{c|}{Phi1.5-1.3B}      & 25.1          & -             & -             & -             & -             & -             & -             & -             & -                      \\
\multicolumn{1}{l|}{Emu3$^{\dagger}$~\cite{wang2024emu3}}             & \multicolumn{1}{c|}{-}                & 31.6          & 47.6          & 58.5          & -             & 68.2          & -             & -             & 70.0*         & 687  \\
\multicolumn{1}{l|}{VILA-U~\cite{wu2024vila}}           & \multicolumn{1}{c|}{LLaMA2-7B}        & -             & -             & -             & -             & 59.0          & -             & -             & -             & -                      \\
\multicolumn{1}{l|}{Janus~\cite{wu2024janus}}            & \multicolumn{1}{c|}{DeepSeek-1B} & 30.5          & 52.2          & 69.4          & -             & 63.7          & -             & -             & -             & -                      \\
\rowcolor[gray]{0.9}
\multicolumn{1}{l|}{UniToken-StageII (Ours)}  & \multicolumn{1}{c|}{Chameleon-7B}     & 34.2          & \textbf{64.1} & 70.3 & 45.2 & \textbf{70.3} & \textbf{41.6} & 49.6 & 67.6 & 634                    \\ 
\rowcolor[gray]{0.9}
\multicolumn{1}{l|}{UniToken-StageIII (Ours)}  & \multicolumn{1}{c|}{Chameleon-7B}     & 32.8          & 62.0 & \textbf{71.1} & \textbf{46.1} & 69.9 & 38.5 & \textbf{57.8} & \textbf{68.7} & \textbf{757}                    \\ 
\bottomrule
\end{tabular}}
\end{center}
\caption{\textbf{Comparison with state-of-the-art methods on prevalent multimodal comprehension benchmarks.} ``\textit{Understanding-Only}" refers to methods that exclusively support the multimodal understanding task, while ``\textit{Understanding \& Generation}" refers to methods that support both multimodal understanding and generation tasks. ``$\dagger$" indicates that Emu3 trains separate models for each of the two tasks individually, and we report the results of its chat version. ``*" indicates that the images of related training datasets are observed during training. ``-" denotes that the results of the approaches are not reported in their paper.}
\label{tab:cmp_und}
\end{table*}

\begin{table*}[!h]
\begin{center}
\scalebox{0.75}{
\begin{tabular}{clcccccccccc}
\toprule
\multicolumn{1}{l|}{}                                                                                       & \multicolumn{1}{c|}{}                & \multicolumn{7}{c|}{GenEval}                                                                                                                                                                                                  & \multicolumn{3}{c}{T2I-CompBench++}                                                   \\
\multicolumn{1}{l|}{Type}                                                                                   & \multicolumn{1}{l|}{Method}          & \multicolumn{1}{l}{Overall} & \multicolumn{1}{l}{Single Obj.} & \multicolumn{1}{l}{Two Obj.} & \multicolumn{1}{l}{Counting} & \multicolumn{1}{l}{Colors} & \multicolumn{1}{l}{Positions} & \multicolumn{1}{l|}{Color Attri.}  & \multicolumn{1}{l}{Color} & \multicolumn{1}{l}{Shape} & \multicolumn{1}{l}{Texture} \\ \midrule
\multicolumn{12}{c}{\textit{Generation-Only}}                                                                                                                                                                                                                                                                                                                                                                                                                                  \\ \midrule
\multicolumn{1}{c|}{\multirow{7}{*}{\textit{Diffusion-based}}}                                                  & \multicolumn{1}{l|}{DALL-E2~\cite{ramesh2022hierarchical}}         & 0.52                        & 0.94                            & 0.66                         & 0.49                         & 0.77                       & 0.10                          & \multicolumn{1}{c|}{0.19}          & 0.5750                    & 0.5464                    & 0.6374                      \\
\multicolumn{1}{c|}{}                                                                                       & \multicolumn{1}{l|}{DALL-E3~\cite{betker2023improving}}         & 0.67                        & 0.96                            & 0.87                         & 0.47                         & 0.83                       & \textbf{0.43}                 & \multicolumn{1}{c|}{0.45}          & \textbf{0.8110}                    & \textbf{0.6750}                    & \textbf{0.8070}                      \\
\multicolumn{1}{c|}{}                                                                                       & \multicolumn{1}{l|}{SDv1.5~\cite{rombach2022high}}          & 0.43                        & 0.97                            & 0.38                         & 0.35                         & 0.76                       & 0.04                          & \multicolumn{1}{c|}{0.06}          & 0.3730                    & 0.3646                    & 0.4219                      \\
\multicolumn{1}{c|}{}                                                                                       & \multicolumn{1}{l|}{SDv2.1~\cite{rombach2022high}}          & 0.50                        & 0.98                            & 0.51                         & 0.44                         & 0.85                       & 0.07                          & \multicolumn{1}{c|}{0.17}          & 0.5694                    & 0.4495                    & 0.4982                      \\
\multicolumn{1}{c|}{}                                                                                       & \multicolumn{1}{l|}{SDXL~\cite{podell2023sdxl}}            & 0.55                        & 0.98                            & 0.74                         & 0.39                         & 0.85                       & 0.15                          & \multicolumn{1}{c|}{0.23}          & 0.6369                    & 0.5408                    & 0.5637                      \\
\multicolumn{1}{c|}{}                                                                                       & \multicolumn{1}{l|}{SD3~\cite
{esser2024scaling}}             & \textbf{0.74}               & \textbf{0.99}                   & \textbf{0.94}                & \textbf{0.72}                & \textbf{0.89}              & 0.33                          & \multicolumn{1}{c|}{\textbf{0.60}} & -                         & -                         & -                           \\
\multicolumn{1}{c|}{}                                                                                       & \multicolumn{1}{l|}{PixArt-$\alpha$~\cite{chen2023pixart}}    & 0.48                        & 0.98                            & 0.50                         & 0.44                         & 0.80                       & 0.08                          & \multicolumn{1}{c|}{0.07}          & 0.6886                    & 0.5582                    & 0.7044                      \\ \hline
\multicolumn{1}{c|}{\textit{AR based}}                                                                      & \multicolumn{1}{l|}{LlamaGen~\cite{sun2024autoregressive}}        & 0.32                        & 0.71                            & 0.34                         & 0.21                         & 0.58                       & 0.07                          & \multicolumn{1}{c|}{0.04}          & -                         & -                         & -                           \\ \midrule
\multicolumn{12}{c}{\textit{Understanding \& Generation}}                                                                                                                                                                                                                                                                                                                                                                                                                               \\ \midrule
\multicolumn{1}{c|}{\multirow{3}{*}{\textit{\begin{tabular}[c]{@{}c@{}}Diffusion \& AR \\ hybrid\end{tabular}}}} & \multicolumn{1}{l|}{Show-o~\cite{xie2024show}}          & 0.53                        & 0.95                            & 0.52                         & \textbf{0.49}                & \textbf{0.82}              & 0.11                          & \multicolumn{1}{c|}{\textbf{0.28}} & -                         & -                         & -                           \\
\multicolumn{1}{c|}{}                                                                                       & \multicolumn{1}{l|}{SEED-X~\cite{ge2024seed}}          & 0.49                        & \textbf{0.97}                   & \textbf{0.58}                & 0.26                         & 0.80                       & \textbf{0.19}                 & \multicolumn{1}{c|}{0.14}          & -                         & -                         & -                           \\
\multicolumn{1}{c|}{}                                                                                       & \multicolumn{1}{l|}{TransFusion~\cite{zhou2024transfusion}}     & \textbf{0.63}               & -                               & -                            & -                            & -                          & -                             & \multicolumn{1}{c|}{-}             & -                         & -                         & -                           \\ \hline
\multicolumn{1}{c|}{\multirow{5}{*}{\textit{AR based}}}                                                     & \multicolumn{1}{l|}{Chameleon~\cite{lu2024chameleon}}       & 0.39                        & -                               & -                            & -                            & -                          & -                             & \multicolumn{1}{c|}{-}             & -                         & -                         & -                           \\
\multicolumn{1}{c|}{}                                                                                       & \multicolumn{1}{l|}{Lumina-mGPT~\cite{liu2024lumina}}            & 0.52                        & \textbf{0.98}                   & 0.72                         & 0.32                & 0.85                       & 0.19                          & \multicolumn{1}{c|}{0.16}          & 0.5558                    & 0.4485                    & 0.5413                      \\
\multicolumn{1}{c|}{}                                                                                       & \multicolumn{1}{l|}{Emu3$^{\dagger}$~\cite{wang2024emu3}}            & 0.54                        & \textbf{0.98}                   & 0.71                         & 0.34                & 0.81                       & 0.17                          & \multicolumn{1}{c|}{0.21}          & 0.6107                    & 0.4734                    & 0.6178                      \\
\multicolumn{1}{c|}{}                                                                                       & \multicolumn{1}{l|}{Janus~\cite{wu2024janus}}           & 0.61               & 0.97                            & 0.68                         & 0.30                         & \textbf{0.84}              & \textbf{0.46}                 & \multicolumn{1}{c|}{\textbf{0.42}} & 0.7552                         & 0.4768                         & 0.6214                           \\
\rowcolor[gray]{0.9}  
\multicolumn{1}{c|}{}                                                  &  \multicolumn{1}{l|}{UniToken-StageII (Ours)} & 0.58                        & 0.98                           & 0.75                & \textbf{0.39}                         & 0.81                       & 0.26                          & \multicolumn{1}{c|}{0.32}          & 0.7115           & 0.5179           & 0.6670             \\
\rowcolor[gray]{0.9}
\multicolumn{1}{c|}{}                                                    &  \multicolumn{1}{l|}{UniToken-StageIII (Ours)} & \textbf{0.63}                        & \textbf{0.99}                            & \textbf{0.80}                & 0.35                         & \textbf{0.84}                       & 0.38                          & \multicolumn{1}{c|}{0.39}          & \textbf{0.7833}           & \textbf{0.5847}           & \textbf{0.7315}             \\
\bottomrule
\end{tabular}}
\end{center}
\caption{\textbf{Comparison with state-of-the-art methods on prevalent multimodal generation benchmarks.} ``\textit{Generation-Only}" refers to methods that exclusively support image generation task. We also categorize methods based on their generation mechanisms, namely diffusion-based, autoregressive-based (AR), andhybrid approaches combining the two. ``$\dagger$" indicates that we use Emu3 generation version for evaluation.  ``-" denotes that the results of the approaches are not reported in their paper.}
\label{tab:cmp_gen}
\end{table*}

\subsection{Inference}
During inference, we employ greedy decoding to ensure deterministic outputs for the visual understanding task, while multinomial sampling decoding is utilized to enhance the diversity of generated images. Additionally, we also employ the classifier-free guidance (CFG) for image generation task following prior works~\cite{wang2024emu3,wu2024janus}, and set the guidance scale as 5.5 and 5.0 when evaluating GenEval~\cite{ghosh2024geneval} and T2i-Compbench++~\cite{huang2023t2i}, respectively.

\begin{table*}[!h]
\begin{center}
\scalebox{0.9}{
\begin{tabular}{l|cc|cc|cc}
\toprule
\multirow{2}{*}{\#} & \multicolumn{2}{c|}{Training Data} & \multicolumn{2}{c|}{Visual Understanding} & \multicolumn{2}{c}{Image Generation} \\ \cmidrule{2-7} 
                    & Visual Und.      & Image Gen.      & General             & TextRead            & GenEval       & T2I-Compbench++      \\ \midrule
1                   & 2M               & 2M              & 49.8                   & 42.9                   & 51.3             & 40.3                    \\
2                   & 2M               & 1M              & 50.0(\blue{$\uparrow$0.2})                & 43.0(\blue{$\uparrow$0.1})                & 50.7(\red{$\downarrow$0.6})          & 39.7(\red{$\downarrow$0.6})                 \\
\cdashline{2-7}[1pt/1pt]
3                   & 10M              & 10M             & 53.3                & 49.5                & 49.4          & 39.4                 \\
4                   & 10M              & 5M              & 53.0(\red{$\downarrow$0.3})                & 49.4(\red{$\downarrow$0.1})                & 38.4(\red{$\downarrow$11.0})          & 33.8(\red{$\downarrow$5.6})                 \\ \bottomrule
\end{tabular}}
\end{center}
\caption{\textbf{Impact of training data scale on the proportion of visual understanding and image generation data.} For robust evaluation, we compute the average scores across MMMU~\cite{yue2024mmmu}, MMB-CN~\cite{liu2024mmbench}, MMB-EN~\cite{liu2024mmbench}, MME~\cite{fu2024mme}, MMStar~\cite{chen2024we}, SEED~\cite{li2023seed}, MathVista~\cite{lu2023mathvista}, and HallusionBench~\cite{guan2023hallusionbench}, with the aggregated result denoted as ``General". Similarly, we calculate the average scores across AI2D~\cite{kembhavi2016diagram}, OCRBench~\cite{liu2023hidden}, and MMVet~\cite{yu2023mm}, with the aggregated result denoted as ``TextRead". For GenEval~\cite{ghosh2024geneval} and T2I-Compbench++~\cite{huang2023t2i}, we report their overall scores.}
\label{tab:data_scale}
\end{table*}

\begin{table*}[!ht]
\begin{center}
\scalebox{0.9}{
\begin{tabular}{l|c|cc|cc|cc}
\toprule
\multirow{2}{*}{\#} & \multirow{2}{*}{Model}     & \multicolumn{2}{c|}{Training Data} & \multicolumn{2}{c|}{Visual Understanding} & \multicolumn{2}{c}{Image Generation} \\ \cmidrule{3-8} 
                    &                            & Visual Und.      & Image Gen.      & General             & TextRead            & GenEval       & T2I-Compbench++      \\ \midrule
1                   & \multirow{3}{*}{Chameleon$^\dagger$} & 10M              & /               & 45.4                & 28.7                & -             & -                    \\
2                   &                            & /                & 10M             & -                   & -                   & 51.2          & 40.1                 \\
3                   &                            & 10M              & 10M             & 45.6(\blue{$\uparrow$0.2})                 & 28.4(\red{$\downarrow$0.3})                & 31.1(\red{$\downarrow$20.1})          & 26.0(\red{$\downarrow$14.1})                 \\
\cdashline{2-8}[1pt/1pt]
4                   & \multirow{3}{*}{UniToken*} & 10M              & /               & 53.2                & 50.1                & -             & -                    \\
5                   &                            & /                & 10M             & -                   & -                   & 51.1          & 39.5                 \\
6                   &                            & 10M              & 10M             & 53.3(\blue{$\uparrow$0.1})                & 49.5(\red{$\downarrow$0.6})                & 49.4(\red{$\downarrow$1.7})          & 39.4(\red{$\downarrow$0.1})                 \\ \bottomrule
\end{tabular}}
\end{center}
\caption{\textbf{Impact of visual encoding approaches on the interference between visual understanding and image generation tasks.} ``$\dagger$" represents that we adopt Chameleon architecture as the discrete-only visual encoding competitor, and train it with our curated dataset for rigorous ablation. ``*" indicates that input resolution scale-up technique is not employed. ``/" signifies that no training data from the corresponding task types is used.  ``-" denotes that the performance metrics are omitted from evaluation, as no data relevant to the corresponding tasks is included in the training set. ``General" and ``TextRead" share the same meaning as in Tab.\ref{tab:data_scale}.}
\label{tab:task_interference}
\end{table*}

\section{Experiments}
In this section, we first detail the implementation specifics of UniToken, encompassing its architectural design and hyper-parameter configurations. Next, we conduct a comprehensive comparison of UniToken against a diverse range of state-of-the-art methods across both visual understanding and image generation benchmarks. Subsequently, we perform ablation studies to dissect the contributions of individual components within UniToken, followed by additional explorations to provide deeper insights. Finally, we present extensive qualitative results to visually demonstrate the capabilities and effectiveness of UniToken.

\begin{figure*}[!h]
  \centering 
  \includegraphics[width=0.95\textwidth]{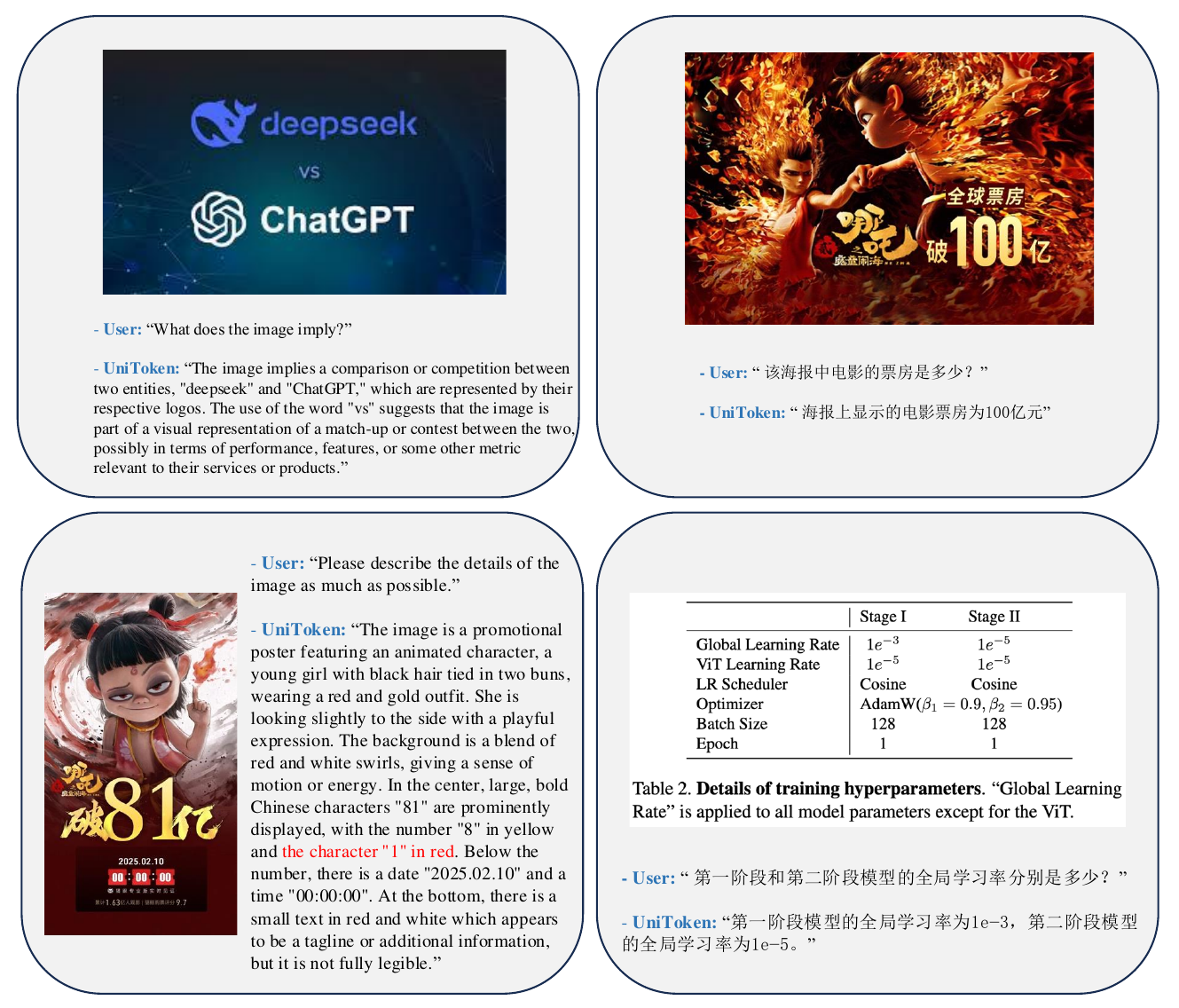} 
  \caption{\textbf{The question answering results of UniToken.} Different types of questions, both in English and Chinese, are evaluated using our UniToken. Hallucinations in the responses are highlighted in \red{red}.} 
  \label{fig:qual_I2T} 
  \vspace{-3mm}
\end{figure*}

\subsection{Implementation Details}
For the LLM component, we adopt the architecture of Chameleon~\cite{lu2024chameleon} and initialize its parameters using the pre-trained checkpoint from Lumina-mGPT~\cite{liu2024lumina}. Therefore, our UniToken inherits the discrete vision tokenizer of Chameleon, which has a codebook of size 16,384 and downsample ratio of 16. For the continuous visual encoder, we utilize SigLIP-SO400M-Patch14-384~\cite{zhai2023sigmoid}. By further applying the aforementioned resolution scale-up technique on the SigLIP, our UniToken achieves a maximum input resolution of $768\times768$. The detailed hyperparameters during the whole training stages are listed in Tab.\ref{tab:hyperparam}. During the experiment, we found that using a larger ViT learning rate (e.g., $5e^{-4}$) causes the SigLIP ViT to collapse severely.

\begin{figure*}[h]
  \centering 
  \includegraphics[width=0.88\textwidth]{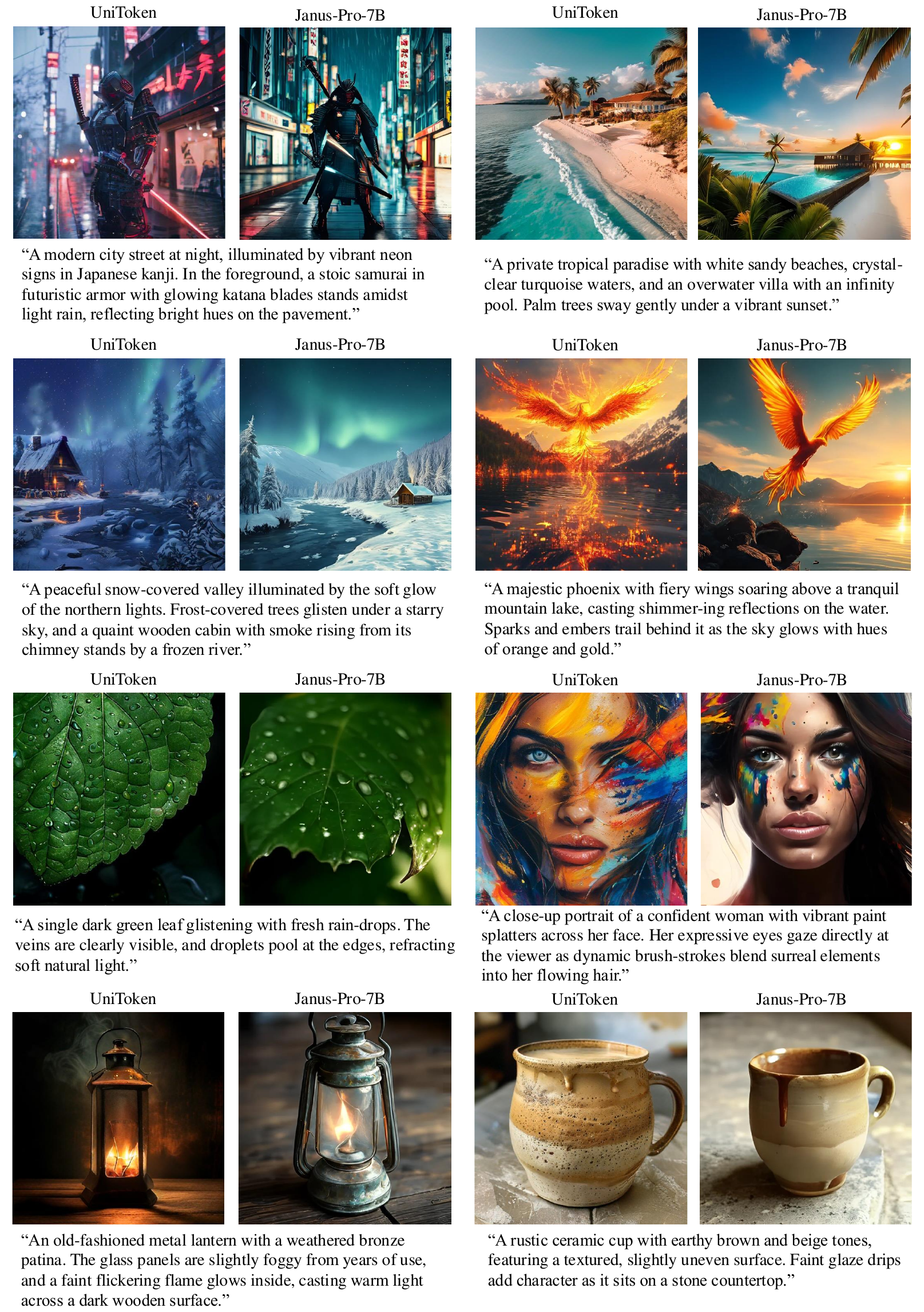} 
  \caption{\textbf{Comparison of image generation results between UniToken and Janus-Pro-7B.}} 
  \label{fig:qual_T2I} 
  \vspace{-3mm}
\end{figure*}

\subsection{Comparison with State-of-the-Arts}
\noindent\textbf{Performances of Multimodal Understanding.}\quad We compare UniToken with MLLMs specialized exclusively for visual understanding tasks and MLLMs capable of both visual understanding and image generation tasks across widely recognized benchmarks, as detailed in Tab.~\ref{tab:cmp_und}. On the one hand, compared with understanding-only specialists, our UniToken achieves better performances across most benchmarks. Notably, even when compared to LLaVA-v1.6(HD), a leading MLLM in the research field, UniToken achieves superior performance by a significant margin (e.g., +5.6 on SEEDBench, +7.0 on MathVista, and +102 on OCRBench). On the other hand, when compared to models capable of unified understanding and generation, UniToken also demonstrates significant performance improvements. It is noteworthy that, despite inheriting the pre-trained checkpoint from Lumina-mGPT, which lacks robust visual understanding capabilities, UniToken still outperforms leading approaches such as Emu3 (chat version) and Janus.  

\noindent\textbf{Performances of Image Generation.}\quad We also compare UniToken with image generation models and approaches with both visual understanding and image generation capabilities as demonstrated in Tab.\ref{tab:cmp_gen}. Since diffusion-based and autoregressive methods are prominent paradigms in image generation, we further categorize approaches according to their generation paradigms. From the Tab.\ref{tab:cmp_gen}, we have the following observations: (1) For image generation models, leading diffusion-based approaches, such as SD3 and DALL-E3, significantly outperforms the autoregressive-based approaches. (2) Although without enjoying the benefits of pre-trained diffusion models, our UniToken achieves competitive or even superior performances when compared with diffusion and autoregressive hybrid approaches. (3) UniToken demonstrates performance comparable to other autoregressive-based unified models. Across specific evaluation metrics, our model achieves higher scores in categories such as ``Two Obj." and ``Counting", but underperforms in ``Positions" and ``Color Attri.". This discrepancy may stem from the absence of relevant text prompts in our text-to-image generation dataset.

\subsection{Comprehensive Analysis}
In this section, we delve into the fundamental design principles of developing a unified model supporting visual understanding and image generation tasks. Specifically, we first examine the interference between these two tasks and then compare the extent of such interference when employing different visual encoding techniques. Subsequently, we investigate the impact of the data proportion between these tasks on model performance and derive insights into selecting appropriate data proportions under varying training dataset scales. For efficiency, we omit Stage III training and the input resolution scale-up technique in all experiments conducted in this section.

\noindent\textbf{Exploration of Task Interference.}\quad In this section, we train a model using three distinct datasets: visual understanding data only, image generation data only, and the combined dataset of visual understanding and image generation data. Meanwhile, to further investigate the effects of visual encoding approaches on task interference, we evaluate two models: (1) Chameleon~\cite{lu2024chameleon}, representing discrete-only visual encoding, and (2) our UniToken, which employs a unified discrete-continuous visual encoding strategy. As illustrated in Tab.\ref{tab:task_interference}, for the Chameleon model, we find that joint training has little effect on the model visual understanding capability (\#3~\emph{vs.}~\#1), but causes severe image generation capability degradation (\#3~\emph{vs.}~\#2). This suggests that the discrete-only visual encoding approach struggles to effectively manage both visual understanding and image generation tasks simultaneously, with the former task tending to dominate the optimization process. On the other hand, our model demonstrates robustness in joint training compared to single-task training, both for visual understanding (\#6~\emph{vs.}~\#4) and image generation (\#6~\emph{vs.}~\#5). This highlights that 
\emph{\textbf{our unified discrete-continuous visual encoding approach is less susceptible to task interference compared to the discrete-only visual encoding method}}.

\noindent\textbf{Exploration of Data Proportion.}\quad
\label{sec_exp:data_distribution}
We ablate the effect of data proportion of visual understanding and image generation tasks under varying data scales as illustrated in Tab.\ref{tab:data_scale}. At a smaller data scale (less than 5M), a 2:1 ratio of visual understanding to image generation data ensures the stability of both capabilities (\#2~\emph{vs.}~\#1). However, when scaling up the training data (more than 15M), a 2:1 ratio leads to significant degradation in image generation performance, whereas adjusting the ratio to 1:1 maintains the effectiveness of both capabilities (\#4~\emph{vs.}~\#3).
We attribute this phenomenon to the fact that, with a 2:1 ratio, \emph{\textbf{scaling up the training data further amplifies the disparity in absolute sample sizes between visual understanding and image generation tasks}}, ultimately resulting in the degradation of image generation performance.


\subsection{Qualitative Results}
\vspace{1.5mm}
\noindent\textbf{Visual Understanding.}\quad For the image-to-text comprehension, we collect images of heating topics from websites and the snapshot of this paper. Afterward, we manually write questions based on the image content. As illustrated in Fig.\ref{fig:qual_I2T}, our UniToken support both English and Chinese question answering, and can tackle diverse question formats to generate corresponding responses. Although some hallucinations are present in these responses, as highlighted in red in Fig.\ref{fig:qual_T2I}, more advanced architectural designs and comprehensive training data aimed at enhancing visual understanding in cutting-edge MLLMs could be incorporated into our UniToken framework in the future.

\vspace{1.5mm}
\noindent\textbf{Image Generation.}\quad For the text-to-image generation, we first automatically generate text prompts using the instruction of \textit{``Please generate some text prompts for generating high-quality images"} to prompt the ChatGPT. With these text prompts, we feed them into our UniToken and the concurrent work Janus-Pro-7B~\cite{chen2025janus}, and demonstrate their results in Fig.\ref{fig:qual_T2I}. Through comparison, it can be observed that the images generated by our UniToken exhibit finer textures and more intricate visual details than those produced by Janus-Pro-7B.


\section{Conclusion}
In this paper, we proposed UniToken, a unified visual representation that seamlessly integrates discrete and continuous tokens to bridge the gap between visual understanding and image generation tasks. Our experiments demonstrate that this approach achieves state-of-the-art performance across diverse multimodal tasks while providing valuable insights into task interference and data distribution challenges. Overall, UniToken establishes a robust foundation for future research in unified multimodal modeling.
{
    \small
    \bibliographystyle{ieeenat_fullname}
    \bibliography{main}
}

\end{document}